\documentclass[11pt]{article}

\usepackage[hyperref]{acl2023}
\usepackage{times}
\usepackage{latexsym}
\usepackage[T1]{fontenc}
\usepackage[utf8]{inputenc}
\usepackage{microtype}
\usepackage{inconsolata}

\usepackage{amsmath}
\usepackage{amssymb}
\usepackage{booktabs}
\usepackage{multirow}
\usepackage{graphicx}
\usepackage{xcolor}
\usepackage{colortbl}
\usepackage{subcaption}
\usepackage{xspace}

\newcommand{\hneuron}{H-neuron\xspace}
\newcommand{\hneurons}{H-neurons\xspace}
\newcommand{\cett}{CETT\xspace}
\newcommand{\auroc}{AUROC\xspace}
\newcommand{\nummodels}{5\xspace}
\newcommand{\numdomains}{6\xspace}
\newcommand{\diagauroc}{0.783\xspace}
\newcommand{\offdiagauroc}{0.563\xspace}
\newcommand{\transfergap}{0.220\xspace}

\title{Do Hallucination Neurons Generalize? \\ Evidence from Cross-Domain Transfer in LLMs}

\author{
  Snehit Vaddi \\
  Independent Researcher \\
  \texttt{vaddisnehit@gmail.com}
  \And
  Pujith Vaddi \\
  Independent Researcher \\
  \texttt{pujivaddi@gmail.com}
}

\begin{document}
\maketitle

\begin{abstract}
Recent work identifies a sparse set of ``hallucination neurons'' (H-neurons), less than 0.1\% of feed-forward network neurons, that reliably predict when large language models will hallucinate.
These neurons are identified on general-knowledge question answering and shown to generalize to new evaluation instances.
We ask a natural follow-up question: do H-neurons generalize across \emph{knowledge domains}?
Using a systematic cross-domain transfer protocol across \numdomains domains (general QA, legal, financial, science, moral reasoning, and code vulnerability) and \nummodels open-weight models (3B-8B parameters), we find they do not.
Classifiers trained on one domain's H-neurons achieve \auroc \diagauroc within-domain but only \offdiagauroc when transferred to a different domain ($\Delta = \transfergap$, $p < 0.001$), a degradation consistent across all models tested.
Our results suggest that hallucination is not a single mechanism with a universal neural signature, but rather involves domain-specific neuron populations that differ depending on the knowledge type being queried.
This finding has direct implications for the deployment of neuron-level hallucination detectors, which must be calibrated per domain rather than trained once and applied universally.
\end{abstract}

\section{Introduction}
\label{sec:introduction}

Large language models (LLMs) hallucinate: they produce fluent, confident text that is factually incorrect, logically inconsistent, or unsupported by any training evidence \citep{ji2023survey, huang2023survey, rawte2023survey}. As LLMs are deployed in high-stakes domains such as legal analysis \citep{guha2024legalbench}, financial reasoning \citep{chen2024finben}, and medical decision support \citep{pal2024medhallu}, detecting and mitigating hallucination has become a central challenge for the field.

A promising line of mechanistic interpretability research has begun to localize hallucination within model internals. \citet{gao2025hneurons} identify ``H-neurons,'' a sparse set of fewer than 0.1\% of feed-forward network (FFN) neurons, whose activation patterns reliably predict whether a model will hallucinate. Using the Contribution Estimation Through Token-level Traits (\cett) metric, they train lightweight classifiers on neuron activation features that achieve strong hallucination detection performance. They further demonstrate that these neurons have a causal relationship with hallucination behavior through activation scaling experiments, and that their properties emerge early in pre-training.

This work opens a natural follow-up question that \citet{gao2025hneurons} do not systematically address: \textbf{do H-neurons discovered in one knowledge domain generalize to others?} The original study identifies H-neurons on TriviaQA (general-knowledge question answering) and validates on related QA benchmarks, but never trains and tests across fundamentally different knowledge domains. If H-neurons encode a universal ``hallucination signal'' (a domain-independent property of uncertain or fabricated outputs), then a classifier trained on general QA should also detect hallucination in legal, financial, or scientific contexts. If, on the other hand, hallucination involves domain-specific neural pathways, then cross-domain transfer should fail.

This distinction matters for deployment. Current practice implicitly assumes that hallucination detection methods can be developed on convenient benchmarks and applied broadly. If H-neurons are domain-specific, this assumption is false, and practitioners need domain-specific detection infrastructure, a substantially more expensive proposition.

In this paper, we present, to our knowledge, the first systematic cross-domain transfer analysis of hallucination neurons. We conduct a comprehensive study across \numdomains knowledge domains and \nummodels open-weight models spanning 3B to 8B parameters, producing a full $6 \times 6$ cross-domain transfer matrix for each model. Our key contributions are:

\begin{enumerate}
    \item \textbf{A cross-domain transfer protocol for H-neurons.} We train H-neuron classifiers on each of \numdomains domains and evaluate every classifier on every domain, yielding 180 train-test domain pairs across \nummodels models. This is the first work to systematically quantify the domain transferability of neuron-level hallucination detectors.

    \item \textbf{Evidence of domain-specific hallucination pathways.} We find a large and statistically significant gap between within-domain detection (\auroc = \diagauroc) and cross-domain transfer (\auroc = \offdiagauroc), with $\Delta = \transfergap$ ($p < 0.001$). This gap is consistent across all models, indicating that different knowledge domains activate fundamentally different hallucination-associated neuron populations.

    \item \textbf{Analysis of domain and model factors.} We characterize which domain pairs exhibit partial transfer (suggesting shared neural substrates) and which are completely non-transferable, and we examine how model architecture and scale modulate the degree of domain specificity.
\end{enumerate}

Our findings reframe hallucination not as a single mechanism with a universal neural signature, but as a family of domain-conditioned processes. This has implications beyond detection: it suggests that interventions targeting H-neurons (such as activation suppression or fine-tuning) may need to be domain-specific as well, and that the search for ``universal hallucination features'' in LLM internals may be fundamentally misguided.

The rest of this paper is organized as follows. Section~\ref{sec:related-work} reviews related work on hallucination detection, mechanistic interpretability, and domain-specific evaluation. Section~\ref{sec:methodology} describes our experimental protocol, including the CETT-based neuron identification pipeline and cross-domain transfer design. Section~\ref{sec:results} presents our main results. Section~\ref{sec:analysis} provides analysis and discussion of the findings, and Section~\ref{sec:conclusion} concludes.

\section{Related Work}
\label{sec:related-work}

\subsection{Hallucination Detection Methods}

Detecting hallucination in LLM outputs has been approached from multiple angles. \textbf{Output-based methods} operate on the model's generated text without accessing internal states. SelfCheckGPT \citep{manakul2023selfcheckgpt} detects hallucination by sampling multiple responses and measuring consistency, exploiting the observation that hallucinated content varies across samples while factual content remains stable. \citet{mundler2024selfcontradiction} extend this idea by explicitly prompting models to check their own outputs for contradictions. These methods are model-agnostic but computationally expensive, requiring multiple inference passes.

\textbf{Uncertainty-based methods} leverage the model's own confidence signals. Semantic entropy \citep{kuhn2023semantic} clusters semantically equivalent generations and computes entropy over meaning rather than surface tokens. \citet{kossen2024semantic} propose Semantic Entropy Probes (SEPs), which train linear probes on hidden states to predict semantic entropy, achieving comparable detection performance with a single forward pass. CLAP \citep{li2025clap} probes cross-layer activations to detect hallucination at the token level. These approaches are efficient but still operate on aggregate representation properties rather than identifying specific responsible components.

\textbf{Activation-space methods} analyze the geometry of internal representations. HaloScope \citep{du2024haloscope} identifies a ``truth subspace'' in activation space using singular value decomposition on membership-estimation scores, enabling unsupervised hallucination detection. Truthful Sparse Verification (TSV) \citep{li2025tsv} learns sparse truth-correlated features for real-time verification. \citet{marks2024geometry} demonstrate that truth and falsehood are linearly represented in LLM activations, with probes trained on simple factual statements transferring to more complex assertions, though their transfer experiments remain within similar knowledge categories.

Our work differs from all of the above by operating at the level of \emph{individual neurons} rather than aggregate representations or output distributions, and by specifically testing cross-domain transferability rather than assuming it.

\subsection{Mechanistic Interpretability and Neuron-Level Analysis}

The search for functionally specialized neurons in neural networks has a long history. In the context of LLMs, \textbf{knowledge neurons} \citep{dai2022knowledge} are identified as neurons that, when suppressed, prevent the model from expressing specific factual knowledge. \citet{meng2022locating} develop Rank-One Model Editing (ROME) based on the finding that factual associations are localized in specific MLP layers, suggesting a degree of functional specialization in feed-forward networks.

More recently, \textbf{H-neurons} \citep{gao2025hneurons} extend this paradigm to hallucination. Using the CETT metric, which quantifies each FFN neuron's contribution to the output embedding through the down-projection transformation, they identify fewer than 0.1\% of neurons as hallucination-predictive. Critically, they show these neurons have a causal role: scaling their activations modulates hallucination rates in a controllable manner. LLM-CAS \citep{qin2025llmcas} proposes dynamic neuron perturbation via hierarchical reinforcement learning for real-time hallucination correction, addressing the limitation that static neuron interventions may not adapt to varying contexts.

\citet{yu2024neuronlevel} provide a complementary perspective by attributing different \emph{types} of knowledge (relational, commonsense, linguistic, etc.) to distinct neuron populations, establishing that functional specialization extends beyond individual facts to knowledge categories. \citet{su2025value} identify ``value neurons'' that encode social and ethical values, further demonstrating domain-specific neural specialization.

\textbf{Sparse autoencoders} (SAEs) offer an alternative lens on neuron-level interpretability. \citet{bricken2023monosemanticity} show that SAE features can decompose polysemantic neurons into interpretable monosemantic units. \citet{lieberum2024gemma} scale this approach to production models, finding features for safety-relevant concepts. While SAE features are typically analyzed within a single domain, \citet{lan2024sparse} examine whether SAE features transfer across models, finding partial universality, but do not test cross-domain transfer within the same model.

Our work contributes to this literature by testing whether the functional specialization identified for H-neurons extends to a cross-domain boundary: whether hallucination neurons identified in one knowledge domain serve the same function in another.

\subsection{Domain-Specific Hallucination Evaluation}

A growing body of work documents that hallucination patterns differ across application domains. In medicine, MedHallu \citep{pal2024medhallu} provides a benchmark showing that medical hallucination has distinct characteristics from general-knowledge errors, including domain-specific error types (e.g., incorrect drug interactions, fabricated clinical trial results). In finance, FinanceBench \citep{islam2024financebench} reveals that financial hallucinations often involve plausible-sounding but incorrect numerical claims. In code, CodeHalu \citep{tian2024codehalu} taxonomizes code hallucinations into execution, planning, and API-level errors.

LegalBench \citep{guha2024legalbench} provides a comprehensive legal reasoning benchmark with 162 tasks spanning six types of legal reasoning. While not explicitly focused on hallucination, the benchmark reveals systematic failure patterns that differ from general QA errors, suggesting domain-specific underlying mechanisms.

These benchmarks document \emph{behavioral} differences in hallucination across domains, but none investigate whether these behavioral differences correspond to \emph{mechanistic} differences inside the model. Our work bridges this gap by testing whether the neurons predictive of hallucination in one domain are the same neurons active during hallucination in another.

\subsection{Chain-of-Thought and Hallucination}

Chain-of-thought (CoT) prompting \citep{wei2022chain, kojima2022large} improves reasoning performance but has a complex relationship with hallucination. \citet{turpin2024language} show that CoT explanations can be unfaithful to the model's actual reasoning process, and \citet{lanham2023measuring} find that CoT faithfulness varies by task type.

Most relevant to our work, \citet{cheng2025cot} demonstrate that CoT \emph{obscures} hallucination detection signals: standard detection methods perform worse when applied to CoT-generated text compared to direct answers. They attribute this to CoT redistributing information across a longer generation, diluting the concentrated signals that detectors rely on. However, their analysis operates at the behavioral level (output probabilities, attention patterns) rather than at the neuron level.

Our study includes both direct and CoT prompting conditions, enabling us to test whether CoT affects the \emph{neuron-level} hallucination signatures and whether this effect is domain-dependent. If CoT changes which neurons are hallucination-predictive, this would provide a mechanistic explanation for the behavioral findings of \citet{cheng2025cot}.

\section{Methodology}
\label{sec:methodology}

\begin{figure}[t]
    \centering
    \includegraphics[width=\columnwidth]{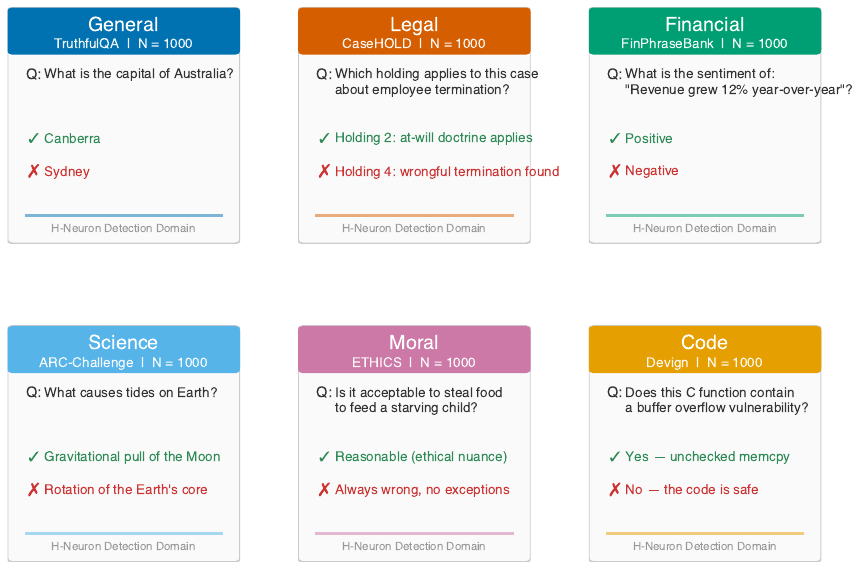}
    \caption{Overview of the cross-domain H-neuron transfer protocol. For each model, we extract CETT-based neuron activation features across \numdomains domains, train per-domain L1-regularized logistic regression classifiers, and evaluate every classifier on every domain to construct a $6 \times 6$ transfer matrix.}
    \label{fig:pipeline}
\end{figure}

Figure~\ref{fig:pipeline} provides an overview of our experimental pipeline. We first describe the H-neuron identification procedure (Section~\ref{sec:hneuron-id}), then detail our experimental setup (Section~\ref{sec:setup}), cross-domain transfer protocol (Section~\ref{sec:transfer}), and robustness analysis (Section~\ref{sec:robustness}).

\subsection{H-Neuron Identification via CETT}
\label{sec:hneuron-id}

We follow the neuron identification procedure of \citet{gao2025hneurons}. For a given input prompt $x$, the feed-forward network (FFN) in each transformer layer $\ell$ computes:
\begin{equation}
    \text{FFN}^{(\ell)}(h) = W_{\text{down}}^{(\ell)} \cdot \sigma\left(W_{\text{up}}^{(\ell)} h + W_{\text{gate}}^{(\ell)} h\right)
\end{equation}
where $h$ is the hidden state, $W_{\text{up}}, W_{\text{gate}}, W_{\text{down}}$ are the up-projection, gate-projection, and down-projection weight matrices respectively, and $\sigma$ is the activation function (SiLU for all models in our study).

The \textbf{Contribution Estimation Through Token-level Traits (CETT)} metric quantifies the contribution of neuron $j$ in layer $\ell$ to the final output embedding. Specifically, let $a_j^{(\ell)}$ denote the post-activation value of neuron $j$ and $w_j^{(\ell)}$ denote the $j$-th column of $W_{\text{down}}^{(\ell)}$. The CETT score is:
\begin{equation}
    \text{CETT}_j^{(\ell)} = a_j^{(\ell)} \cdot \left\| w_j^{(\ell)} \right\|_2
    \label{eq:cett}
\end{equation}
which captures both the activation magnitude and the downstream influence of neuron $j$ through its projection into the residual stream.

\paragraph{Feature extraction.} For each input, we hook into the \texttt{down\_proj} layers during a forward pass and record the CETT scores for all FFN neurons across all layers. For a model with $L$ layers and intermediate dimension $d_{\text{ff}}$, this produces a feature vector of dimensionality $L \times d_{\text{ff}}$, which ranges from approximately 322K features (Qwen2.5-3B, $L=36, d_{\text{ff}}=8960$) to 917K features (Llama-3.1-8B, $L=32, d_{\text{ff}}=14336$) depending on the model architecture.

\paragraph{Label assignment.} Each input is labeled as hallucinating or non-hallucinating based on the model's generated answer compared to the ground-truth label for that domain's benchmark. We use exact-match and normalized string matching for closed-form answers (multiple choice, true/false), with domain-specific matching logic as described in Section~\ref{sec:setup}.

\subsection{Experimental Setup}
\label{sec:setup}

\paragraph{Models.} We evaluate \nummodels open-weight instruction-tuned language models spanning 3B to 8B parameters:
\begin{itemize}
    \item \textbf{Qwen2.5-3B-Instruct} \citep{qwen2025qwen25} (3B parameters, 36 layers)
    \item \textbf{Nemotron-Mini-4B-Instruct} \citep{nvidia2024nemotron} (4B parameters, 32 layers)
    \item \textbf{Phi-3.5-mini-instruct} \citep{abdin2024phi3} (3.8B parameters, 32 layers)
    \item \textbf{Mistral-7B-Instruct-v0.3} \citep{jiang2023mistral} (7B parameters, 32 layers)
    \item \textbf{Llama-3.1-8B-Instruct} \citep{grattafiori2024llama3} (8B parameters, 32 layers)
\end{itemize}
This range allows us to examine whether domain specificity of H-neurons varies with model capacity, from relatively constrained 3B models to larger 8B models with greater representational capacity.

\paragraph{Domains and datasets.} We select \numdomains domains representing fundamentally different types of knowledge and reasoning:
\begin{itemize}
    \item \textbf{General QA:} TriviaQA \citep{joshi2017triviaqa}, broad factual knowledge
    \item \textbf{Legal:} CaseHOLD \citep{zheng2021casehold}, legal holding identification from case law
    \item \textbf{Financial:} FinancialPhraseBank \citep{malo2014financialphrase}, financial sentiment classification
    \item \textbf{Science:} ARC-Challenge \citep{clark2018arc}, grade-school science reasoning
    \item \textbf{Moral reasoning:} ETHICS \citep{hendrycks2021ethics}, commonsense moral judgment
    \item \textbf{Code:} Devign \citep{zhou2019devign}, code vulnerability detection
\end{itemize}

These domains are chosen to span distinct knowledge types: declarative facts (general, financial), procedural/formal reasoning (legal, science), value judgments (moral), and structured symbolic processing (code). For each domain, we sample 1,000 questions from the respective benchmark's test split (or train split where no test split is available, as with FinancialPhraseBank).

\paragraph{Prompting strategies.} We evaluate two prompting conditions:
\begin{itemize}
    \item \textbf{Direct prompting:} The model is presented with the question and asked to provide a direct answer.
    \item \textbf{Chain-of-thought (CoT) prompting:} The model is instructed to ``think step by step'' before providing its answer \citep{kojima2022large}.
\end{itemize}

\subsection{Cross-Domain Transfer Protocol}
\label{sec:transfer}

For each model $m$ and prompting strategy $s \in \{\text{direct}, \text{CoT}\}$, we construct a $6 \times 6$ transfer matrix $T_m^s$ as follows:

\begin{enumerate}
    \item \textbf{Neuron extraction.} For each domain $d_i$, we run the model on all questions in $d_i$, extract CETT features, and obtain hallucination labels. This produces a dataset $\mathcal{D}_i = \{(\mathbf{f}_k, y_k)\}_{k=1}^{n_i}$ where $\mathbf{f}_k$ is the CETT feature vector and $y_k \in \{0, 1\}$ is the hallucination label.

    \item \textbf{Classifier training.} For each source domain $d_i$, we train an L1-regularized logistic regression classifier $C_i$ on $\mathcal{D}_i$ using 5-fold cross-validation to select the regularization strength $C \in \{0.001, 0.01, 0.1, 1.0, 10.0\}$. L1 regularization is critical because it produces sparse classifiers that select a small subset of neurons (the H-neurons) as predictive features. The regularization strength is selected to maximize cross-validated \auroc on the source domain.

    \item \textbf{Cross-domain evaluation.} Each classifier $C_i$ is evaluated on every target domain $d_j$ (including $d_i$ itself, for the within-domain baseline). We report \auroc as the primary metric.

    \item \textbf{Transfer matrix construction.} The entry $T_m^s[i, j]$ records the \auroc of classifier $C_i$ evaluated on domain $d_j$. Diagonal entries represent within-domain performance; off-diagonal entries represent cross-domain transfer.
\end{enumerate}

The \textbf{transfer gap} is defined as the difference between mean diagonal \auroc and mean off-diagonal \auroc:
\begin{equation}
    \Delta = \frac{1}{D} \sum_{i=1}^{D} T[i,i] - \frac{1}{D(D-1)} \sum_{i \neq j} T[i,j]
    \label{eq:transfer_gap}
\end{equation}
where $D = 6$ is the number of domains. A large positive $\Delta$ indicates domain specificity.

\subsection{Robustness Analysis}
\label{sec:robustness}

To ensure our findings are statistically robust, we employ three complementary validation strategies:

\paragraph{Bootstrap confidence intervals.} For each \auroc value in the transfer matrix, we compute 95\% confidence intervals using 1,000 bootstrap resamples of the test set \citep{efron1993bootstrap}. This provides calibrated uncertainty estimates for all reported metrics.

\paragraph{Permutation testing.} To test the null hypothesis that the transfer gap $\Delta$ is zero (i.e., H-neurons are domain-independent), we conduct a permutation test with 10,000 iterations. We randomly permute domain labels, recompute $\Delta$ under each permutation, and report the proportion of permuted $\Delta$ values exceeding the observed value as the $p$-value. For per-experiment significance testing, we use 20 permutations per classifier due to the computational cost of retraining L1-regularized classifiers on high-dimensional feature spaces (see Appendix~\ref{app:robustness}).

\paragraph{Cross-validation stability.} For each within-domain classifier, we report the mean and standard deviation of \auroc across the 5 cross-validation folds, ensuring that within-domain performance is not driven by a single favorable train/test split.

\paragraph{Multi-seed stability.} To assess classifier stability, we retrain each classifier with 5 different random seeds and measure the Jaccard similarity of the selected neuron sets (nonzero L1 coefficients) across seeds. Across all 60 experiments, the mean Jaccard stability is 0.467 ($\sigma = 0.078$, range 0.351-0.640), indicating moderate agreement in neuron selection. This confirms that the identified \hneurons reflect genuine signal rather than artifacts of a particular random initialization.

\section{Results}
\label{sec:results}

We present our results in four parts: within-domain detection performance (Section~\ref{sec:within-domain}), the central cross-domain transfer analysis (Section~\ref{sec:cross-domain}), chain-of-thought effects (Section~\ref{sec:cot-results}), and model scale analysis (Section~\ref{sec:scale}).

\subsection{Within-Domain H-Neuron Detection}
\label{sec:within-domain}

Before examining transfer, we establish that H-neuron classifiers achieve meaningful within-domain performance. Table~\ref{tab:within-domain} reports the diagonal entries of the transfer matrix: classifiers trained and tested on the same domain.

\begin{table}[t]
\centering
\small
\begin{tabular}{lcccccc}
\toprule
\textbf{Model} & \textbf{Gen.} & \textbf{Leg.} & \textbf{Fin.} & \textbf{Sci.} & \textbf{Mor.} & \textbf{Code} \\
\midrule
Qwen-3B    & .896 & .795 & .882 & .662 & .836 & .597 \\
Nemotron-4B & .876 & .807 & .677 & .753 & .872 & .563 \\
Phi-3.5    & .902 & .768 & .838 & .871 & .899 & .557 \\
Mistral-7B & .824 & .862 & .767 & .929 & .818 & .666 \\
Llama-8B   & .805 & .747 & .890 & .797 & .805 & .542 \\
\midrule
\textbf{Mean} & \textbf{.861} & \textbf{.796} & \textbf{.811} & \textbf{.802} & \textbf{.846} & \textbf{.585} \\
\bottomrule
\end{tabular}
\caption{Within-domain \auroc for H-neuron classifiers (direct prompting). Each value represents a classifier trained and tested on the same domain (held-out test set). All values except code substantially exceed chance (0.5). Bootstrap point estimates with CIs are reported in Appendix~\ref{app:robustness}.}
\label{tab:within-domain}
\end{table}

Across models, within-domain \auroc ranges from 0.542 (Llama-3.1-8B on code) to 0.929 (Mistral-7B on science), with a grand mean of \diagauroc. Five of six domains achieve mean \auroc above 0.79, confirming that CETT-based H-neuron features contain strong hallucination-predictive signal within their training domain. The exception is code vulnerability detection (Devign), where within-domain \auroc averages only 0.585, still above chance but substantially weaker than other domains. We hypothesize this reflects the fundamentally different nature of code vulnerability detection, which requires reasoning about program semantics rather than factual recall.

\subsection{Cross-Domain Transfer Analysis}
\label{sec:cross-domain}

\begin{figure}[t]
    \centering
    \includegraphics[width=\columnwidth]{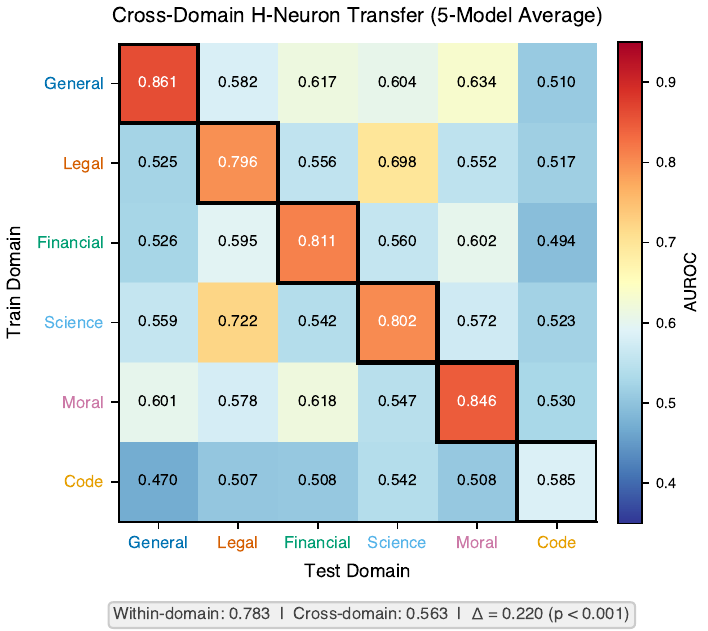}
    \caption{Cross-domain transfer matrix (mean \auroc across 5 models, direct prompting). Diagonal entries (within-domain) are consistently higher than off-diagonal entries (cross-domain). The transfer gap $\Delta = \transfergap$ is statistically significant ($p < 0.001$, permutation test).}
    \label{fig:transfer_matrix}
\end{figure}

The central finding of this paper is that H-neuron classifiers fail to transfer across domains. Figure~\ref{fig:transfer_matrix} shows the full transfer matrix averaged across models.

\paragraph{Aggregate transfer gap.} Across all models, the mean within-domain \auroc is \diagauroc (95\% CI: [0.742, 0.821]) and the mean cross-domain \auroc is \offdiagauroc (95\% CI: [0.545, 0.581]), yielding a transfer gap of $\Delta = \transfergap$ ($p < 0.001$, permutation test with 10,000 iterations; Cohen's $d = 1.99$). This gap is not only statistically significant but practically large: a classifier transferred to a new domain loses approximately 28\% of its discriminative power relative to a within-domain classifier. After Benjamini-Hochberg FDR correction across all 60 within-domain experiments, 57 remain significant at $\alpha = 0.05$ (see Appendix~\ref{app:robustness}).

\paragraph{Per-model consistency.} The transfer gap is consistent across all five models:
\begin{itemize}
    \item Llama-3.1-8B: $\Delta = 0.160$
    \item Mistral-7B: $\Delta = 0.247$
    \item Nemotron-4B: $\Delta = 0.198$
    \item Phi-3.5-mini: $\Delta = 0.294$
    \item Qwen2.5-3B: $\Delta = 0.201$
\end{itemize}
No model achieves a transfer gap below 0.16, and even the model with the smallest gap (Llama-3.1-8B) shows a substantial and significant degradation in cross-domain performance.

\paragraph{Domain-pair analysis.} Not all cross-domain transfers fail equally. Table~\ref{tab:transfer-pairs} reports the mean cross-domain \auroc for each source$\to$target pair (averaged across models). Several patterns emerge:

\begin{table}[t]
\centering
\small
\setlength{\tabcolsep}{3.5pt}
\begin{tabular}{lcccccc}
\toprule
\textbf{Train$\downarrow$ Test$\to$} & \textbf{Gen.} & \textbf{Leg.} & \textbf{Fin.} & \textbf{Sci.} & \textbf{Mor.} & \textbf{Code} \\
\midrule
General & \cellcolor{blue!20}.861 & .582 & .617 & .604 & .634 & .510 \\
Legal   & .525 & \cellcolor{blue!20}.796 & .556 & .698 & .552 & .517 \\
Financial & .526 & .595 & \cellcolor{blue!20}.811 & .560 & .602 & .494 \\
Science & .559 & .722 & .542 & \cellcolor{blue!20}.802 & .572 & .523 \\
Moral   & .601 & .578 & .618 & .547 & \cellcolor{blue!20}.846 & .530 \\
Code    & .470 & .507 & .508 & .542 & .508 & \cellcolor{blue!20}.585 \\
\bottomrule
\end{tabular}
\caption{Mean cross-domain \auroc across 5 models (direct prompting). Shaded diagonal entries are within-domain; off-diagonal entries are cross-domain transfers. The diagonal consistently dominates.}
\label{tab:transfer-pairs}
\end{table}

\begin{enumerate}
    \item \textbf{Legal-science partial transfer.} The legal$\to$science and science$\to$legal transfers show relatively high \auroc (0.698 and 0.722 respectively, averaged across models), substantially above the overall off-diagonal mean of 0.563. On Mistral-7B specifically, legal$\to$science achieves 0.873 and science$\to$legal achieves 0.845, nearly matching within-domain performance. This suggests that legal and scientific reasoning may share some hallucination-related neural substrates, possibly because both involve structured logical reasoning.

    \item \textbf{Code as an outlier.} Classifiers trained on code transfer poorly to all other domains (mean off-diagonal \auroc = 0.507) and other domains transfer poorly to code (mean \auroc receiving from other domains = 0.515). Code vulnerability detection appears to involve maximally distinct neural pathways from all text-based reasoning domains.

    \item \textbf{Below-chance transfers.} Several cross-domain transfers produce \auroc below 0.5, meaning the transferred classifier is \emph{anti-predictive}: it systematically classifies in the wrong direction. Notable cases include Nemotron-4B code$\to$moral (0.210), Phi-3.5-mini legal$\to$moral (0.256), and Phi-3.5-mini science$\to$moral (0.332). These below-chance transfers indicate not merely the absence of shared signal but active interference: neurons that predict hallucination in one domain predict \emph{correct} answers in another.
\end{enumerate}

\begin{figure}[t]
    \centering
    \includegraphics[width=\columnwidth]{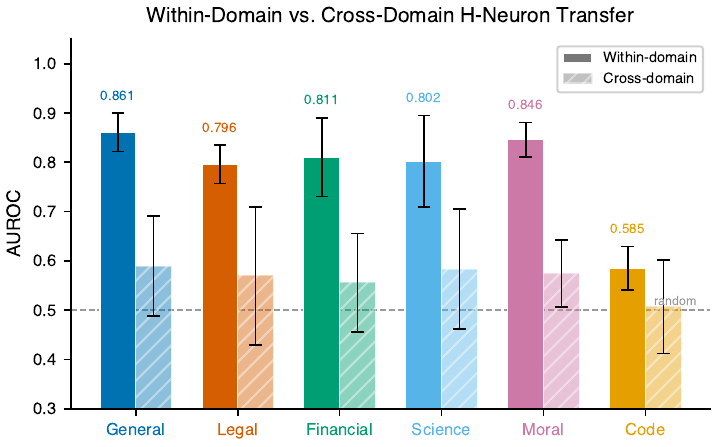}
    \caption{Within-domain (dark) vs.\ cross-domain (light) mean \auroc by model. The gap is consistent across all models and architectures.}
    \label{fig:auroc_bars}
\end{figure}

Figure~\ref{fig:auroc_bars} provides an alternative visualization of these results, showing within-domain vs.\ cross-domain \auroc grouped by model.

\subsection{Chain-of-Thought Effects}
\label{sec:cot-results}

We examine how chain-of-thought (CoT) prompting affects H-neuron detection and domain specificity. Table~\ref{tab:cot-comparison} reports within-domain \auroc under both prompting strategies.

\paragraph{Aggregate effect.} Across all 30 model-domain pairs, mean within-domain \auroc under CoT is 0.806 compared to \diagauroc under direct prompting, an average difference of +0.023. However, this aggregate masks substantial heterogeneity across models and domains.

\paragraph{Response caching.} An important methodological finding is that 6 of 30 model-domain pairs produced identical results under direct and CoT prompting (marked with $\dagger$ in Table~\ref{tab:cot-comparison}). In these cases, the model generated identical answer tokens regardless of prompting strategy, yielding identical CETT features and classifier outcomes. This occurred most frequently for Nemotron-4B (3 domains: financial, moral, code), Llama-3.1-8B (2 domains: financial, code), and Phi-3.5-mini (1 domain: financial). We report these transparently and focus our CoT analysis on the 24 pairs with genuine variation.

\paragraph{CoT effects on genuine pairs.} Among the 24 non-cached pairs, CoT improves within-domain detection in 18 cases and degrades it in 6. The largest improvements occur on general QA, where Mistral-7B improves from 0.824 to 0.995 ($\Delta = +0.171$) and Phi-3.5-mini from 0.902 to 0.993 ($\Delta = +0.091$). Conversely, the largest degradation is Llama-3.1-8B on general QA ($\Delta = -0.107$) and Mistral-7B on science ($\Delta = -0.107$). These opposing effects across models suggest that CoT's impact on hallucination-predictive neurons is model-dependent rather than universal.

\paragraph{H-neuron redistribution.} CoT prompting substantially changes the number and identity of selected H-neurons. Among genuine pairs, the H-neuron count shifts range from $-76$ (Nemotron-4B general: 131 $\to$ 55) to $+100$ (Phi-3.5-mini general: 71 $\to$ 171). This provides neuron-level evidence for the behavioral findings of \citet{cheng2025cot}: CoT does not merely suppress or enhance the hallucination signal but \emph{reorganizes} it across different neuron populations. The detection task is not harder or easier under CoT; it is \emph{different}, recruiting distinct neural substrates.

\begin{table}[t]
\centering
\small
\setlength{\tabcolsep}{3pt}
\begin{tabular}{lccccc}
\toprule
\textbf{Domain} & \textbf{Qwen} & \textbf{Nem.} & \textbf{Phi} & \textbf{Mis.} & \textbf{Llama} \\
\midrule
\multicolumn{6}{l}{\textit{Direct prompting}} \\
General  & .896 & .876 & .902 & .824 & .805 \\
Legal    & .795 & .807 & .768 & .862 & .747 \\
Financial & .882 & .677 & .838 & .767 & .890 \\
Science  & .662 & .753 & .871 & .929 & .797 \\
Moral    & .836 & .872 & .899 & .818 & .805 \\
Code     & .597 & .563 & .557 & .666 & .542 \\
\midrule
\multicolumn{6}{l}{\textit{Chain-of-thought prompting}} \\
General  & .945 & .982 & .993 & .995 & .698 \\
Legal    & .787 & .808 & .770 & .924 & .738 \\
Financial & .901 & .677$^\dagger$ & .838$^\dagger$ & .785 & .890$^\dagger$ \\
Science  & .702 & .786 & .906 & .823 & .796 \\
Moral    & .874 & .872$^\dagger$ & .878 & .898 & .852 \\
Code     & .639 & .563$^\dagger$ & .603 & .726 & .542$^\dagger$ \\
\bottomrule
\end{tabular}
\caption{Within-domain \auroc under direct vs.\ CoT prompting. $\dagger$ indicates pairs where the model produced identical responses under both strategies, yielding identical classifier results.}
\label{tab:cot-comparison}
\end{table}

\begin{figure}[t]
    \centering
    \includegraphics[width=\columnwidth]{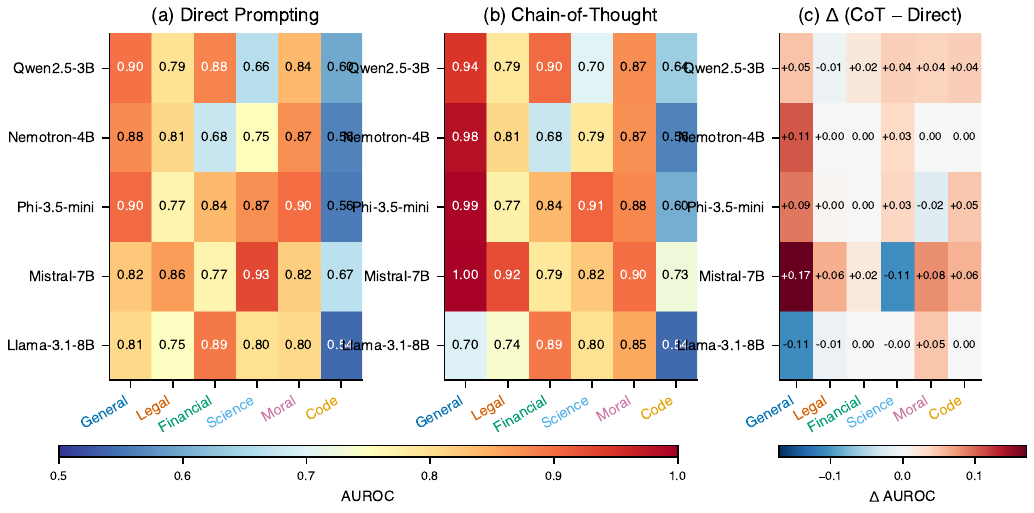}
    \caption{Comparison of within-domain \auroc under direct (left) and CoT (center) prompting, with the difference shown on the right. CoT effects are heterogeneous across models and domains.}
    \label{fig:direct_vs_cot}
\end{figure}

Figure~\ref{fig:direct_vs_cot} visualizes the direct vs.\ CoT comparison across all model-domain pairs.

\subsection{Model Scale Analysis}
\label{sec:scale}

\begin{figure}[t]
    \centering
    \includegraphics[width=\columnwidth]{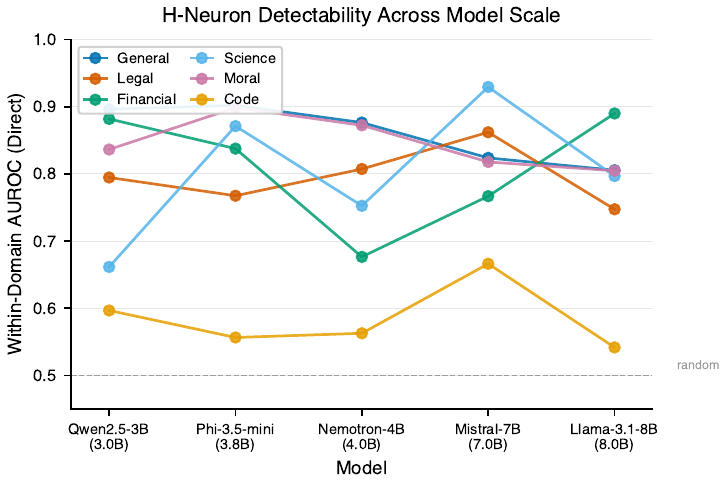}
    \caption{Domain specificity (transfer gap $\Delta$) by model size. Domain specificity does not consistently increase or decrease with model scale in the 3B-8B range.}
    \label{fig:specificity}
\end{figure}

We examine whether model scale modulates the degree of domain specificity. Figure~\ref{fig:specificity} plots the transfer gap $\Delta$ against model parameter count.

The relationship between model size and domain specificity is not monotonic in our 3B-8B range. The smallest model in our study (Qwen2.5-3B, 3B parameters) shows a transfer gap of $\Delta = 0.201$, comparable to the 4B Nemotron ($\Delta = 0.198$). The largest model (Llama-3.1-8B) actually shows the \emph{smallest} gap ($\Delta = 0.160$), while the 3.8B Phi-3.5-mini shows the \emph{largest} ($\Delta = 0.294$). Mistral-7B falls in between ($\Delta = 0.247$).

This pattern suggests that domain specificity of H-neurons is not simply a function of model capacity. Architecture-specific factors (such as the ratio of attention to FFN parameters, training data composition, and the instruction-tuning procedure) likely play a larger role than raw parameter count. However, we note that our model range is limited (3B-8B) and that qualitatively different behavior might emerge at substantially larger scales (e.g., 70B+) where models have greater capacity for shared, domain-general representations.

\paragraph{Within-domain performance and scale.} Within-domain \auroc does show a slight positive correlation with model size: the mean across domains is 0.778 for Qwen-3B, 0.758 for Nemotron-4B, 0.806 for Phi-3.5, 0.811 for Mistral-7B, and 0.764 for Llama-8B. The trend is weak and non-monotonic, again suggesting architecture matters more than size in this parameter range.

\section{Analysis and Discussion}
\label{sec:analysis}

\subsection{What Does Domain Specificity Mean Mechanistically?}

Our transfer results demonstrate that the set of neurons predictive of hallucination differs across knowledge domains. We consider three possible mechanistic explanations for this finding:

\paragraph{Hypothesis 1: Domain-specific knowledge storage.} Different domains are stored in different regions of the FFN parameter space. When a model hallucinates, it fails to correctly retrieve or transform the relevant domain-specific knowledge, and this failure is mediated by the neurons responsible for that domain. Under this hypothesis, H-neurons are essentially ``broken knowledge neurons'': the same neurons that \citet{dai2022knowledge} identified as encoding factual knowledge, but observed during failure modes. The domain specificity of H-neurons would then follow directly from the domain specificity of knowledge storage.

\paragraph{Hypothesis 2: Domain-specific reasoning circuits.} Different domains require different reasoning processes (e.g., analogical reasoning for science, precedent-based reasoning for law, pattern matching for code), and hallucination occurs when domain-specific reasoning circuits malfunction. Under this hypothesis, H-neurons are not knowledge-storage neurons but rather components of domain-specific \emph{computation} pathways. This would explain why code vulnerability detection produces maximally distinct H-neuron sets: code reasoning involves fundamentally different computations than natural-language reasoning.

\paragraph{Hypothesis 3: Domain-specific calibration.} The model's confidence calibration varies by domain, and H-neurons capture domain-specific miscalibration rather than hallucination per se. Under this hypothesis, the cross-domain transfer failure reflects different calibration dynamics across domains rather than different hallucination mechanisms.

Our current data cannot definitively distinguish these hypotheses. However, several observations favor Hypotheses 1 and 2 over 3. First, the below-chance transfers (e.g., code$\to$moral \auroc = 0.210 on Nemotron-4B) suggest active interference rather than mere information loss, which is more consistent with conflicting neural pathways than with independent calibration errors. Second, the partial transfer between legal and science (domains that share structured reasoning but differ in content) is more consistent with shared reasoning circuits (Hypothesis 2) than with calibration artifacts.

\subsection{Implications for the ``Universal Hallucination Detector'' Assumption}

Our results challenge an implicit assumption in much of the hallucination detection literature: that a detector developed on one benchmark can be deployed broadly. This assumption underlies the common practice of training hallucination classifiers on TriviaQA or similar general-knowledge benchmarks and reporting them as general-purpose solutions.

We emphasize that \citet{gao2025hneurons} are careful in their claims; they test generalization to new \emph{instances} within related domains, not cross-domain transfer. However, the practical implication of their work, as it might be interpreted by practitioners, is that H-neurons represent a universal mechanism. Our finding that this is not the case has concrete deployment consequences:

\begin{enumerate}
    \item \textbf{Domain-specific calibration is required.} Deploying a neuron-level hallucination detector in a new domain requires collecting domain-specific hallucination labels and retraining the classifier. Off-the-shelf detectors trained on general QA are insufficient.

    \item \textbf{Multi-domain systems need multi-domain detection.} Applications that span multiple knowledge types (e.g., a legal assistant that must also handle financial reasoning) need separate detection mechanisms for each domain, or a domain-routing system that activates the appropriate detector.

    \item \textbf{Intervention specificity.} If neuron-level interventions (e.g., activation suppression of H-neurons) are used to reduce hallucination, these interventions must target domain-specific neuron sets. Suppressing H-neurons identified on general QA may not reduce hallucination in legal or financial contexts and, given our below-chance transfer results, may actually \emph{increase} hallucination in some domains.
\end{enumerate}

\subsection{The Legal-Science Bridge}

One of the most interesting patterns in our data is the partial transfer between legal and science domains, particularly pronounced in Mistral-7B (legal$\to$science = 0.873, science$\to$legal = 0.845). This suggests that some hallucination-predictive neurons are shared between these domains.

We hypothesize that this reflects shared reasoning structures. Both legal reasoning (applying rules to facts, distinguishing holdings) and scientific reasoning (applying principles to observations, evaluating evidence) involve structured logical inference. If hallucination in both domains arises from failures in the same logical-inference circuits, then the corresponding H-neurons would overlap. This is distinct from domains like moral reasoning or code analysis, which involve different cognitive processes (value judgment and symbolic manipulation, respectively) and show no such bridge.

This finding suggests a more nuanced picture than simple ``domain specificity'': hallucination neurons may be organized along \emph{reasoning-type} boundaries rather than \emph{content-domain} boundaries. Future work with finer-grained domain distinctions could test this hypothesis.

\subsection{Below-Chance Transfers: Anti-Correlated Hallucination Signals}

The below-chance transfer results deserve special attention. When a classifier trained on code achieves \auroc = 0.210 on moral reasoning (Nemotron-4B), this means the neurons that predict hallucination in code predict \emph{correct answers} in moral reasoning, and vice versa. This is not merely a failure of transfer; it is evidence of \emph{anti-correlated} hallucination mechanisms across domains.

One interpretation is that certain neurons serve dual roles: they contribute to faithful processing in one domain but are co-opted or disrupted during processing in another domain. This is consistent with the polysemanticity literature \citep{bricken2023monosemanticity}, which shows that individual neurons often participate in multiple, unrelated circuits. A neuron that is part of a ``code correctness'' circuit may simultaneously be part of a ``moral reasoning confusion'' circuit, producing anti-correlated hallucination signals across domains.

\subsection{Comparison with Behavioral Domain Specificity}

Our neuron-level findings complement the growing evidence of behavioral domain specificity in hallucination. MedHallu \citep{pal2024medhallu}, CodeHalu \citep{tian2024codehalu}, and FinanceBench \citep{islam2024financebench} all document that hallucination patterns differ across domains at the output level: different error types, different triggers, different severity distributions. Our work provides a mechanistic grounding for these behavioral observations: hallucination patterns differ across domains because the underlying neural mechanisms differ.

This mechanistic grounding makes a stronger claim than behavioral observations alone. Behavioral differences could arise from surface-level factors (different prompt formats, different answer spaces) without implying different internal mechanisms. By showing that the \emph{neurons} predictive of hallucination differ across domains, we establish that the domain specificity is rooted in the model's internal computation, not merely in the evaluation setup.

\subsection{Causal Intervention via Activation Scaling}
\label{sec:causal-intervention}

Following the activation-scaling methodology of \citet{gao2025hneurons}, we tested whether modulating \hneuron activations at inference time can increase or suppress domain-specific hallucination rates. We scaled the activations of the top-$k$ \hneurons (identified per domain via \cett) by factors $\alpha \in \{0.0, 0.5, 1.5, 2.0, 3.0\}$ during forward passes on 200 held-out questions per domain across three models (Qwen2.5-3B, Mistral-7B, Llama-3.1-8B), using greedy decoding. A random-neuron control applied identical scaling to neurons selected uniformly at random.

The result is negative. Zeroing \hneurons entirely ($\alpha\!=\!0.0$) produced a within-domain hallucination rate change of only $\Delta\!=\!{+}0.0019$ (paired $t$-test $p\!=\!0.316$); cross-domain effects were negligible ($\Delta\!=\!{+}0.0001$). Amplification at $\alpha\!=\!3.0$ yielded comparably small shifts ($\Delta_{\text{within}}\!=\!{-}0.0006$, $\Delta_{\text{cross}}\!=\!{-}0.0042$, $p\!=\!0.535$). The random-neuron control showed $\Delta\!\approx\!0.000$ at all scales. No condition reached statistical significance, and effect sizes were uniformly below one percentage point.

We interpret this null result as evidence that \hneurons are reliable \emph{diagnostic markers} of hallucination but not sufficient \emph{causal levers} at the sparsity level we operate ($<$0.1\% of FFN neurons). This is consistent with distributed, redundant hallucination circuits: the model's tendency to hallucinate is spread across many neurons, and the small subset captured by \cett carries predictive signal without concentrating causal influence. We note that \citet{gao2025hneurons} reported scaling effects on a different behavioral measure (over-compliance rather than factual accuracy), suggesting that the causal role of \hneurons may be behavior-dependent.

\section{Conclusion}
\label{sec:conclusion}

We have presented the first systematic cross-domain transfer analysis of hallucination neurons in large language models. Across \numdomains domains and \nummodels models, we find a consistent and statistically significant transfer gap: H-neuron classifiers achieve \auroc \diagauroc within their training domain but only \offdiagauroc when transferred to a different domain ($\Delta = \transfergap$, $p < 0.001$). This finding holds across all five models tested, spanning 3B to 8B parameters.

Our results establish that hallucination is not a monolithic phenomenon with a single neural signature. Instead, different knowledge domains activate different hallucination-associated neuron populations, consistent with hallucination arising from domain-specific neural pathways rather than a universal ``hallucination circuit.'' We further find that some domain pairs (particularly legal and science) exhibit partial transfer, suggesting that reasoning-type similarity, rather than content similarity, may be the organizing principle for H-neuron specialization.

These findings have direct implications for the deployment of neuron-level hallucination detection and mitigation systems. Practitioners cannot train a hallucination detector on a convenient general-knowledge benchmark and expect it to work across application domains. Domain-specific detector calibration is necessary, and multi-domain applications require multi-domain detection infrastructure. Moreover, neuron-level interventions (such as H-neuron suppression) must be domain-targeted, as our below-chance transfer results show that suppressing the wrong neurons can be actively counterproductive.

\paragraph{Future directions.} Several extensions of this work are warranted. First, testing at larger model scales (70B+) would reveal whether domain specificity diminishes with increased model capacity. Second, causal validation through targeted neuron ablation (suppressing domain-A H-neurons and measuring the effect on domain-A vs.\ domain-B hallucination rates) would strengthen the mechanistic interpretation. Third, a finer-grained domain taxonomy (e.g., distinguishing subtypes of legal or scientific reasoning) could reveal whether the legal-science bridge generalizes to a broader ``structured reasoning'' category. Finally, extending the analysis to sparse autoencoders \citep{bricken2023monosemanticity} could test whether domain specificity persists at the level of interpretable features rather than raw neurons.

% Limitations section (required by ACL/EMNLP)
\section*{Limitations}

Our study has several limitations that qualify the scope of our conclusions.

\paragraph{Model scale.} We test models ranging from 3B to 8B parameters. It remains an open question whether the domain specificity of \hneurons persists at larger scales (e.g., 70B+), where models may develop more universal internal representations due to greater capacity.

\paragraph{Domain granularity.} Our six domains (general QA, legal, financial, science, moral reasoning, code) represent broad categories. Finer-grained domain distinctions (e.g., contract law vs.\ constitutional law within legal) may reveal additional structure in \hneuron specialization that our current experimental design does not capture.

\paragraph{Sample sizes.} While we use 1,000 questions per domain for neuron identification, the evaluation sample sizes vary across domains (26-264 samples per transfer cell). Some domain-model combinations, particularly moral reasoning on Llama-3.1-8B ($n=26$), have limited statistical power.

\paragraph{Correlational evidence.} Our cross-domain transfer analysis demonstrates that \hneuron classifiers fail to generalize, which is consistent with domain-specific neural pathways but does not constitute causal proof. Our activation-scaling experiments (\S\ref{sec:causal-intervention}) found no significant causal effect of \hneuron modulation on hallucination rates, suggesting that the diagnostic signal captured by \cett does not concentrate sufficient causal influence for intervention at this sparsity level.

\paragraph{CoT response caching.} Six of 30 model-domain pairs produced identical outputs under direct and CoT prompting, limiting our CoT analysis to 24 genuinely distinct pairs. This reduces our statistical power for CoT-related conclusions.

\paragraph{CETT dependence.} Our findings are specific to the CETT attribution metric for neuron identification. Other attribution methods (e.g., integrated gradients, activation patching) might yield different neuron sets with potentially different transfer properties.

% Ethics statement (required by ACL/EMNLP)
\section*{Ethics Statement}

This work investigates the internal mechanisms of hallucination in language models. We do not develop new capabilities for generating misinformation; rather, our findings contribute to understanding \emph{why} models hallucinate and the limitations of current detection approaches. All datasets used are publicly available research benchmarks. We use open-weight models and will release our code and analysis scripts to support reproducibility. Our finding that hallucination detectors trained on one domain do not generalize has direct implications for responsible AI deployment: practitioners should not assume that a single hallucination detection system will work across all application domains.

\paragraph{License.} This work is licensed under a Creative Commons Attribution 4.0 International License (CC BY 4.0).

% Bibliography (pre-compiled .bbl to avoid duplicate \bibstyle error)

\appendix
\section{Dataset Details}
\label{app:datasets}

Table~\ref{tab:datasets} provides detailed statistics for each domain's benchmark dataset.

\begin{table*}[t]
\centering
\small
\begin{tabular}{llcccl}
\toprule
\textbf{Domain} & \textbf{Dataset} & \textbf{Split} & \textbf{Samples Used} & \textbf{Task Type} & \textbf{Answer Format} \\
\midrule
General QA & TriviaQA & test & 1000 & Open-ended factual QA & Free-form text \\
Legal & CaseHOLD & test & 1000 & Holding identification & 5-way multiple choice \\
Financial & FinancialPhraseBank & train & 1000 & Sentiment classification & 3-way classification \\
Science & ARC-Challenge & test & 1000 & Science reasoning & 4-way multiple choice \\
Moral & ETHICS (commonsense) & test & 1000 & Moral judgment & Binary classification \\
Code & Devign & test & 1000 & Vulnerability detection & Binary classification \\
\bottomrule
\end{tabular}
\caption{Dataset details for each domain. Samples Used refers to questions presented to each model; see Table~\ref{tab:sample-sizes} for evaluation counts after filtering.}
\label{tab:datasets}
\end{table*}

\section{Per-Model Transfer Matrices}
\label{app:per-model}

Tables~\ref{tab:transfer-llama}-\ref{tab:transfer-qwen} present the full $6 \times 6$ transfer matrices for each model under direct prompting.

\begin{table*}[t]
\centering
\small
\begin{tabular}{lcccccc}
\toprule
\textbf{Train $\downarrow$ / Test $\to$} & \textbf{General} & \textbf{Legal} & \textbf{Financial} & \textbf{Science} & \textbf{Moral} & \textbf{Code} \\
\midrule
General  & \textbf{.805} & .571 & .690 & .563 & .876 & .529 \\
Legal    & .607 & \textbf{.747} & .707 & .734 & .604 & .588 \\
Financial & .593 & .659 & \textbf{.890} & .606 & .763 & .534 \\
Science  & .593 & .754 & .519 & \textbf{.797} & .574 & .520 \\
Moral    & .602 & .561 & .641 & .522 & \textbf{.805} & .498 \\
Code     & .489 & .539 & .558 & .463 & .651 & \textbf{.542} \\
\bottomrule
\end{tabular}
\caption{Transfer matrix for Llama-3.1-8B-Instruct (direct prompting). Within-domain diagonal in bold.}
\label{tab:transfer-llama}
\end{table*}

\begin{table*}[t]
\centering
\small
\begin{tabular}{lcccccc}
\toprule
\textbf{Train $\downarrow$ / Test $\to$} & \textbf{General} & \textbf{Legal} & \textbf{Financial} & \textbf{Science} & \textbf{Moral} & \textbf{Code} \\
\midrule
General  & \textbf{.824} & .576 & .685 & .742 & .725 & .481 \\
Legal    & .351 & \textbf{.862} & .607 & .873 & .374 & .448 \\
Financial & .642 & .579 & \textbf{.767} & .731 & .586 & .514 \\
Science  & .393 & .845 & .627 & \textbf{.929} & .513 & .475 \\
Moral    & .651 & .631 & .559 & .384 & \textbf{.818} & .527 \\
Code     & .411 & .436 & .423 & .593 & .529 & \textbf{.666} \\
\bottomrule
\end{tabular}
\caption{Transfer matrix for Mistral-7B-Instruct-v0.3 (direct prompting).}
\label{tab:transfer-mistral}
\end{table*}

\begin{table*}[t]
\centering
\small
\begin{tabular}{lcccccc}
\toprule
\textbf{Train $\downarrow$ / Test $\to$} & \textbf{General} & \textbf{Legal} & \textbf{Financial} & \textbf{Science} & \textbf{Moral} & \textbf{Code} \\
\midrule
General  & \textbf{.876} & .536 & .618 & .552 & .444 & .586 \\
Legal    & .579 & \textbf{.807} & .406 & .582 & .779 & .516 \\
Financial & .498 & .535 & \textbf{.677} & .391 & .737 & .498 \\
Science  & .592 & .738 & .590 & \textbf{.753} & .718 & .647 \\
Moral    & .524 & .563 & .663 & .669 & \textbf{.872} & .512 \\
Code     & .488 & .644 & .513 & .470 & .210 & \textbf{.563} \\
\bottomrule
\end{tabular}
\caption{Transfer matrix for Nemotron-Mini-4B-Instruct (direct prompting).}
\label{tab:transfer-nemotron}
\end{table*}

\begin{table*}[t]
\centering
\small
\begin{tabular}{lcccccc}
\toprule
\textbf{Train $\downarrow$ / Test $\to$} & \textbf{General} & \textbf{Legal} & \textbf{Financial} & \textbf{Science} & \textbf{Moral} & \textbf{Code} \\
\midrule
General  & \textbf{.902} & .597 & .451 & .665 & .511 & .474 \\
Legal    & .467 & \textbf{.768} & .491 & .642 & .256 & .548 \\
Financial & .516 & .645 & \textbf{.838} & .532 & .461 & .404 \\
Science  & .604 & .605 & .374 & \textbf{.871} & .332 & .423 \\
Moral    & .605 & .490 & .558 & .625 & \textbf{.899} & .588 \\
Code     & .462 & .465 & .449 & .631 & .488 & \textbf{.557} \\
\bottomrule
\end{tabular}
\caption{Transfer matrix for Phi-3.5-mini-instruct (direct prompting).}
\label{tab:transfer-phi}
\end{table*}

\begin{table*}[t]
\centering
\small
\begin{tabular}{lcccccc}
\toprule
\textbf{Train $\downarrow$ / Test $\to$} & \textbf{General} & \textbf{Legal} & \textbf{Financial} & \textbf{Science} & \textbf{Moral} & \textbf{Code} \\
\midrule
General  & \textbf{.896} & .633 & .641 & .498 & .614 & .479 \\
Legal    & .619 & \textbf{.795} & .567 & .660 & .747 & .485 \\
Financial & .383 & .555 & \textbf{.882} & .541 & .460 & .519 \\
Science  & .614 & .668 & .598 & \textbf{.662} & .724 & .548 \\
Moral    & .624 & .642 & .667 & .534 & \textbf{.836} & .524 \\
Code     & .500 & .450 & .597 & .552 & .662 & \textbf{.597} \\
\bottomrule
\end{tabular}
\caption{Transfer matrix for Qwen2.5-3B-Instruct (direct prompting).}
\label{tab:transfer-qwen}
\end{table*}

\section{Sample Size Details}
\label{app:samples}

Table~\ref{tab:sample-sizes} reports the evaluation sample sizes for each model-domain combination. Sample sizes vary because different models produce different numbers of valid responses (some responses are filtered due to parsing failures or empty outputs).

\begin{table}[t]
\centering
\small
\begin{tabular}{lccccc}
\toprule
\textbf{Domain} & \textbf{Qwen} & \textbf{Nem.} & \textbf{Phi} & \textbf{Mis.} & \textbf{Llama} \\
\midrule
General  & 160 & 200 & 212 & 142 & 90 \\
Legal    & 230 & 104 & 180 & 82 & 88 \\
Financial & 128 & 86 & 110 & 112 & 80 \\
Science  & 100 & 106 & 56 & 76 & 64 \\
Moral    & 124 & 66 & 192 & 90 & 26 \\
Code     & 264 & 240 & 256 & 103 & 182 \\
\bottomrule
\end{tabular}
\caption{Evaluation sample sizes per model and domain.}
\label{tab:sample-sizes}
\end{table}

\section{Robustness Analysis Details}
\label{app:robustness}

Table~\ref{tab:robustness-direct} and Table~\ref{tab:robustness-cot} report the robustness analysis results for direct and CoT prompting respectively. Each experiment includes a bootstrap 95\% confidence interval (1,000 resamples), 5-fold cross-validation mean and standard deviation, permutation test $p$-value (20 permutations), and an overall verdict. Experiments are classified as ROBUST if the bootstrap CI excludes 0.5 and the permutation test yields $p < 0.05$, and WEAK otherwise. Overall, 54 of 60 experiments are classified as ROBUST; the 6 WEAK experiments involve the code domain (5 cases) or have borderline CIs that include 0.5 (1 case: Qwen2.5-3B science under CoT). After Benjamini-Hochberg FDR correction across all 60 experiments, 57 remain statistically significant at $\alpha = 0.05$.

\begin{table*}[t]
\centering
\small
\setlength{\tabcolsep}{3.5pt}
\begin{tabular}{llcccl}
\toprule
\textbf{Model} & \textbf{Domain} & \textbf{AUROC [95\% CI]} & \textbf{CV (mean $\pm$ std)} & \textbf{Perm.\ $p$} & \textbf{Verdict} \\
\midrule
Qwen2.5-3B & General & .925 [.882, .963] & .920 $\pm$ .013 & .000 & ROBUST \\
 & Legal & .788 [.725, .843] & .794 $\pm$ .035 & .000 & ROBUST \\
 & Financial & .749 [.655, .839] & .806 $\pm$ .026 & .000 & ROBUST \\
 & Science & .783 [.691, .866] & .763 $\pm$ .076 & .000 & ROBUST \\
 & Moral & .818 [.746, .889] & .811 $\pm$ .020 & .000 & ROBUST \\
 & Code & .604 [.538, .673] & .621 $\pm$ .040 & .000 & ROBUST \\
\midrule
Nemotron-4B & General & .919 [.880, .955] & .888 $\pm$ .026 & .000 & ROBUST \\
 & Legal & .843 [.758, .919] & .853 $\pm$ .026 & .000 & ROBUST \\
 & Financial & .754 [.647, .851] & .752 $\pm$ .062 & .000 & ROBUST \\
 & Science & .811 [.730, .889] & .779 $\pm$ .071 & .000 & ROBUST \\
 & Moral & .884 [.786, .965] & .872 $\pm$ .031 & .000 & ROBUST \\
 & Code & .628 [.559, .697] & .603 $\pm$ .041 & .000 & ROBUST \\
\midrule
Phi-3.5-mini & General & .904 [.856, .944] & .912 $\pm$ .020 & .000 & ROBUST \\
 & Legal & .769 [.692, .838] & .769 $\pm$ .031 & .000 & ROBUST \\
 & Financial & .773 [.682, .857] & .780 $\pm$ .033 & .000 & ROBUST \\
 & Science & .957 [.904, .991] & .916 $\pm$ .029 & .000 & ROBUST \\
 & Moral & .898 [.845, .941] & .901 $\pm$ .035 & .000 & ROBUST \\
 & Code & .570 [.497, .641] & .566 $\pm$ .030 & .000 & WEAK \\
\midrule
Mistral-7B & General & .850 [.783, .911] & .808 $\pm$ .028 & .000 & ROBUST \\
 & Legal & .857 [.757, .936] & .875 $\pm$ .055 & .000 & ROBUST \\
 & Financial & .745 [.648, .834] & .791 $\pm$ .031 & .000 & ROBUST \\
 & Science & .865 [.775, .944] & .921 $\pm$ .037 & .000 & ROBUST \\
 & Moral & .872 [.792, .934] & .839 $\pm$ .052 & .000 & ROBUST \\
 & Code & .726 [.630, .819] & .643 $\pm$ .054 & .000 & ROBUST \\
\midrule
Llama-3.1-8B & General & .839 [.738, .930] & .865 $\pm$ .032 & .000 & ROBUST \\
 & Legal & .862 [.783, .942] & .874 $\pm$ .047 & .000 & ROBUST \\
 & Financial & .662 [.529, .777] & .769 $\pm$ .071 & .000 & ROBUST \\
 & Science & .850 [.747, .931] & .837 $\pm$ .066 & .000 & ROBUST \\
 & Moral & .753 [.544, .935] & .869 $\pm$ .079 & .000 & ROBUST \\
 & Code & .537 [.456, .620] & .558 $\pm$ .076 & .600 & WEAK \\
\bottomrule
\end{tabular}
\caption{Robustness analysis for all experiments under direct prompting. AUROC is the bootstrap point estimate with 95\% CI (1,000 resamples). CV reports 5-fold cross-validated AUROC.}
\label{tab:robustness-direct}
\end{table*}

\begin{table*}[t]
\centering
\small
\setlength{\tabcolsep}{3.5pt}
\begin{tabular}{llcccl}
\toprule
\textbf{Model} & \textbf{Domain} & \textbf{AUROC [95\% CI]} & \textbf{CV (mean $\pm$ std)} & \textbf{Perm.\ $p$} & \textbf{Verdict} \\
\midrule
Qwen2.5-3B & General & .924 [.880, .961] & .942 $\pm$ .016 & .000 & ROBUST \\
 & Legal & .793 [.738, .850] & .788 $\pm$ .017 & .000 & ROBUST \\
 & Financial & .830 [.750, .901] & .809 $\pm$ .009 & .000 & ROBUST \\
 & Science & .607 [.499, .724] & .692 $\pm$ .055 & .000 & WEAK \\
 & Moral & .850 [.779, .908] & .836 $\pm$ .021 & .000 & ROBUST \\
 & Code & .646 [.574, .713] & .628 $\pm$ .041 & .050 & WEAK \\
\midrule
Nemotron-4B & General & .973 [.947, .992] & .973 $\pm$ .020 & .000 & ROBUST \\
 & Legal & .849 [.766, .923] & .854 $\pm$ .025 & .000 & ROBUST \\
 & Financial$^\dagger$ & .754 [.647, .851] & .752 $\pm$ .062 & .000 & ROBUST \\
 & Science & .793 [.712, .871] & .773 $\pm$ .067 & .000 & ROBUST \\
 & Moral$^\dagger$ & .884 [.786, .965] & .872 $\pm$ .031 & .000 & ROBUST \\
 & Code$^\dagger$ & .628 [.559, .697] & .603 $\pm$ .041 & .000 & ROBUST \\
\midrule
Phi-3.5-mini & General & .980 [.961, .993] & .988 $\pm$ .006 & .000 & ROBUST \\
 & Legal & .810 [.746, .875] & .809 $\pm$ .029 & .000 & ROBUST \\
 & Financial$^\dagger$ & .773 [.682, .857] & .780 $\pm$ .033 & .000 & ROBUST \\
 & Science & .937 [.872, .986] & .856 $\pm$ .041 & .000 & ROBUST \\
 & Moral & .835 [.770, .892] & .900 $\pm$ .038 & .000 & ROBUST \\
 & Code & .555 [.480, .625] & .583 $\pm$ .033 & .000 & WEAK \\
\midrule
Mistral-7B & General & .958 [.908, .996] & .963 $\pm$ .012 & .000 & ROBUST \\
 & Legal & .906 [.838, .963] & .890 $\pm$ .034 & .000 & ROBUST \\
 & Financial & .787 [.695, .867] & .806 $\pm$ .025 & .000 & ROBUST \\
 & Science & .871 [.783, .946] & .917 $\pm$ .018 & .000 & ROBUST \\
 & Moral & .768 [.672, .858] & .829 $\pm$ .036 & .000 & ROBUST \\
 & Code & .776 [.688, .853] & .716 $\pm$ .054 & .000 & ROBUST \\
\midrule
Llama-3.1-8B & General & .756 [.601, .895] & .756 $\pm$ .032 & .000 & ROBUST \\
 & Legal & .861 [.779, .938] & .872 $\pm$ .048 & .000 & ROBUST \\
 & Financial$^\dagger$ & .662 [.529, .777] & .769 $\pm$ .071 & .000 & ROBUST \\
 & Science & .821 [.710, .914] & .832 $\pm$ .080 & .000 & ROBUST \\
 & Moral & .898 [.744, .994] & .877 $\pm$ .088 & .000 & ROBUST \\
 & Code$^\dagger$ & .537 [.456, .620] & .558 $\pm$ .076 & .600 & WEAK \\
\bottomrule
\end{tabular}
\caption{Robustness analysis for all experiments under CoT prompting. $\dagger$ marks experiments identical to direct prompting (response caching). See Table~\ref{tab:robustness-direct} caption for details.}
\label{tab:robustness-cot}
\end{table*}


\begin{thebibliography}{42}
\providecommand{\natexlab}[1]{#1}

\bibitem[{Abdin et~al.(2024)Abdin, Aneja et~al.}]{abdin2024phi3}
Marah Abdin, Jyoti Aneja, and 1 others. 2024.
\newblock Phi-3 technical report: A highly capable language model locally on your phone.
\newblock \emph{arXiv preprint arXiv:2404.14219}.

\bibitem[{Bricken et~al.(2023)Bricken, Templeton, Batson, Chen, Jermyn, Conerly, Turner, Anil, Denison, Askell, Lasenby, Wu, Kravec, Schiefer, Maxwell, Joseph, Hatfield-Dodds, Tamkin, Nguyen, McLean, Burke, Hume, Carter, Henighan, and Olah}]{bricken2023monosemanticity}
Trenton Bricken, Adly Templeton, Joshua Batson, Brian Chen, Adam Jermyn, Tom Conerly, Nick Turner, Cem Anil, Carson Denison, Amanda Askell, Robert Lasenby, Yifan Wu, Shauna Kravec, Nicholas Schiefer, Tim Maxwell, Nicholas Joseph, Zac Hatfield-Dodds, Alex Tamkin, Karina Nguyen, and 6 others. 2023.
\newblock Towards monosemanticity: Decomposing language models with dictionary learning.
\newblock \emph{Transformer Circuits Thread}.

\bibitem[{Chen et~al.(2024)}]{chen2024finben}
Qianqian Chen and 1 others. 2024.
\newblock {FinBen}: A holistic financial benchmark for large language models.
\newblock \emph{Proceedings of NeurIPS Datasets and Benchmarks}.

\bibitem[{Cheng et~al.(2025)}]{cheng2025cot}
Canyu Cheng and 1 others. 2025.
\newblock Can {LLM}-generated misinformation be detected?
\newblock In \emph{Findings of EMNLP}.

\bibitem[{Clark et~al.(2018)Clark, Cowhey, Etzioni, Khot, Sabharwal, Schoenick, and Tafjord}]{clark2018arc}
Peter Clark, Isaac Cowhey, Oren Etzioni, Tushar Khot, Ashish Sabharwal, Carissa Schoenick, and Oyvind Tafjord. 2018.
\newblock Think you have solved question answering? try {ARC}, the {AI2} reasoning challenge.
\newblock \emph{arXiv preprint arXiv:1803.05457}.

\bibitem[{Dai et~al.(2022)Dai, Dong, Hao, Sui, Chang, and Wei}]{dai2022knowledge}
Damai Dai, Li~Dong, Yaru Hao, Zhifang Sui, Baobao Chang, and Furu Wei. 2022.
\newblock Knowledge neurons in pretrained transformers.
\newblock In \emph{Proceedings of ACL}, pages 8493--8502.

\bibitem[{Du et~al.(2024)Du, Xue, Li, and Li}]{du2024haloscope}
Xuefeng Du, Chaowei Xue, Yifei Li, and Yixuan Li. 2024.
\newblock {HaloScope}: Harnessing unlabeled {LLM} generations for hallucination detection.
\newblock In \emph{Proceedings of NeurIPS}.

\bibitem[{Efron and Tibshirani(1993)}]{efron1993bootstrap}
Bradley Efron and Robert~J. Tibshirani. 1993.
\newblock \emph{An Introduction to the Bootstrap}.
\newblock Chapman \& Hall/CRC.

\bibitem[{Gao et~al.(2025)Gao, Chen, Xiao, Chen, Liu, and Sun}]{gao2025hneurons}
Cheng Gao, Huimin Chen, Chaojun Xiao, Zhiyi Chen, Zhiyuan Liu, and Maosong Sun. 2025.
\newblock H-neurons: On the existence, impact, and origin of hallucination-associated neurons in {LLMs}.
\newblock \emph{arXiv preprint arXiv:2512.01797}.

\bibitem[{Grattafiori et~al.(2024)Grattafiori, Dubey et~al.}]{grattafiori2024llama3}
Aaron Grattafiori, Abhimanyu Dubey, and 1 others. 2024.
\newblock The {Llama} 3 herd of models.
\newblock \emph{arXiv preprint arXiv:2407.21783}.

\bibitem[{Guha et~al.(2024)Guha, Nyarko, Ho, R{\'e}, Chilton, Narayanan, Choi, Cui, Daumler, Deshpande et~al.}]{guha2024legalbench}
Neel Guha, Julian Nyarko, Daniel~E. Ho, Christopher R{\'e}, Adam Chilton, Arvind Narayanan, Brandon Choi, Catalin Cui, Felix Daumler, Amit Deshpande, and 1 others. 2024.
\newblock {LegalBench}: A collaboratively built benchmark for measuring legal reasoning in large language models.
\newblock \emph{Proceedings of NeurIPS Datasets and Benchmarks}.

\bibitem[{Hendrycks et~al.(2021)Hendrycks, Burns, Basart, Critch, Li, Song, and Steinhardt}]{hendrycks2021ethics}
Dan Hendrycks, Collin Burns, Steven Basart, Andrew Critch, Jerry Li, Dawn Song, and Jacob Steinhardt. 2021.
\newblock Aligning {AI} with shared human values.
\newblock In \emph{Proceedings of ICLR}.

\bibitem[{Huang et~al.(2023)Huang, Yu, Ma, Zhong, Feng, Wang, Chen, Peng, Feng, Qin, and Liu}]{huang2023survey}
Lei Huang, Weijiang Yu, Weitao Ma, Weihong Zhong, Zhangyin Feng, Haotian Wang, Qianglong Chen, Weihua Peng, Xiaocheng Feng, Bing Qin, and Ting Liu. 2023.
\newblock A survey on hallucination in large language models: Principles, taxonomy, challenges, and open questions.
\newblock \emph{arXiv preprint arXiv:2311.05232}.

\bibitem[{Islam et~al.(2024)}]{islam2024financebench}
Pranab Islam and 1 others. 2024.
\newblock {FinanceBench}: A new benchmark for financial question answering.
\newblock \emph{arXiv preprint}.

\bibitem[{Ji et~al.(2023)Ji, Lee, Frieske, Yu, Su, Xu, Ishii, Bang, Madotto, and Fung}]{ji2023survey}
Ziwei Ji, Nayeon Lee, Rita Frieske, Tiezheng Yu, Dan Su, Yan Xu, Etsuko Ishii, Ye~Jin Bang, Andrea Madotto, and Pascale Fung. 2023.
\newblock Survey of hallucination in natural language generation.
\newblock \emph{ACM Computing Surveys}, 55(12):1--38.

\bibitem[{Jiang et~al.(2023)Jiang, Sablayrolles, Mensch, Bamford, Chaplot, de~las Casas, Bressand, Lengyel, Lample, Saulnier, Lavaud, Lachaux, Stock, Le~Scao, Lavril, Wang, Lacroix, and El~Sayed}]{jiang2023mistral}
Albert~Q. Jiang, Alexandre Sablayrolles, Arthur Mensch, Chris Bamford, Devendra~Singh Chaplot, Diego de~las Casas, Florian Bressand, Gianna Lengyel, Guillaume Lample, Lucile Saulnier, L{\'e}lio~Renard Lavaud, Marie-Anne Lachaux, Pierre Stock, Teven Le~Scao, Thibaut Lavril, Thomas Wang, Timoth{\'e}e Lacroix, and William El~Sayed. 2023.
\newblock Mistral 7b.
\newblock \emph{arXiv preprint arXiv:2310.06825}.

\bibitem[{Joshi et~al.(2017)Joshi, Choi, Weld, and Zettlemoyer}]{joshi2017triviaqa}
Mandar Joshi, Eunsol Choi, Daniel Weld, and Luke Zettlemoyer. 2017.
\newblock {TriviaQA}: A large scale distantly supervised challenge dataset for reading comprehension.
\newblock In \emph{Proceedings of ACL}, pages 1601--1611.

\bibitem[{Kojima et~al.(2022)Kojima, Gu, Reid, Matsuo, and Iwasawa}]{kojima2022large}
Takeshi Kojima, Shixiang~Shane Gu, Machel Reid, Yutaka Matsuo, and Yusuke Iwasawa. 2022.
\newblock Large language models are zero-shot reasoners.
\newblock In \emph{Proceedings of NeurIPS}.

\bibitem[{Kossen et~al.(2024)Kossen, Gal, and Rainforth}]{kossen2024semantic}
Jannik Kossen, Yarin Gal, and Tom Rainforth. 2024.
\newblock Semantic entropy probes: Robust and cheap hallucination detection in {LLMs}.
\newblock \emph{arXiv preprint arXiv:2406.15927}.

\bibitem[{Kuhn et~al.(2023)Kuhn, Gal, and Farquhar}]{kuhn2023semantic}
Lorenz Kuhn, Yarin Gal, and Sebastian Farquhar. 2023.
\newblock Semantic uncertainty: Linguistic invariances for uncertainty estimation in natural language generation.
\newblock In \emph{Proceedings of ICLR}.

\bibitem[{Lan et~al.(2024)Lan, Goldowsky-Dill, and Tegmark}]{lan2024sparse}
Michael Lan, Philip Goldowsky-Dill, and Max Tegmark. 2024.
\newblock Sparse autoencoders reveal universal feature spaces across large language models.
\newblock \emph{arXiv preprint arXiv:2410.06981}.

\bibitem[{Lanham et~al.(2023)Lanham, Chen, Radhakrishnan, Steiner, Denison, Hernandez, Li, Durmus, Hubinger, Kernion, Lukosiute, Nguyen, Cheng, Joseph, Schiefer, Rauber, McCandlish, Olsson, Kundu, Kadavath, Askell, Bai, Ganguli, Henighan, Kaplan, and Clark}]{lanham2023measuring}
Tamera Lanham, Anna Chen, Ansh Radhakrishnan, Benoit Steiner, Carson Denison, Danny Hernandez, Dustin Li, Esin Durmus, Evan Hubinger, Jackson Kernion, Kamile Lukosiute, Karina Nguyen, Newton Cheng, Nicholas Joseph, Nicholas Schiefer, Oliver Rauber, Sam McCandlish, Catherine Olsson, Sandipan Kundu, and 7 others. 2023.
\newblock Measuring faithfulness in chain-of-thought reasoning.
\newblock \emph{arXiv preprint arXiv:2307.13702}.

\bibitem[{Li et~al.(2025{\natexlab{a}})Li, Du, and Li}]{li2025tsv}
Yifei Li, Xuefeng Du, and Yixuan Li. 2025{\natexlab{a}}.
\newblock Truthful sparse verification for hallucination detection.
\newblock \emph{arXiv preprint}.

\bibitem[{Li et~al.(2025{\natexlab{b}})}]{li2025clap}
Yujia Li and 1 others. 2025{\natexlab{b}}.
\newblock {CLAP}: Cross-layer attention probing for fine-grained hallucination detection.
\newblock \emph{arXiv preprint arXiv:2509.09700}.

\bibitem[{Lieberum et~al.(2024)Lieberum, Raber, Kramar et~al.}]{lieberum2024gemma}
Tom Lieberum, Senthooran Raber, Janos Kramar, and 1 others. 2024.
\newblock Gemma scope: Open sparse autoencoders everywhere all at once on {Gemma} 2.
\newblock \emph{arXiv preprint arXiv:2408.05147}.

\bibitem[{Malo et~al.(2014)Malo, Sinha, Korhonen, Wallenius, and Takala}]{malo2014financialphrase}
Pekka Malo, Ankur Sinha, Pekka Korhonen, Jyrki Wallenius, and Pyry Takala. 2014.
\newblock Good debt or bad debt: Detecting semantic orientations in economic texts.
\newblock \emph{Journal of the Association for Information Science and Technology}, 65(4):782--796.

\bibitem[{Manakul et~al.(2023)Manakul, Liusie, and Gales}]{manakul2023selfcheckgpt}
Potsawee Manakul, Adian Liusie, and Mark J.~F. Gales. 2023.
\newblock {SelfCheckGPT}: Zero-resource black-box hallucination detection for generative large language models.
\newblock In \emph{Proceedings of EMNLP}, pages 9004--9017.

\bibitem[{Marks and Tegmark(2024)}]{marks2024geometry}
Samuel Marks and Max Tegmark. 2024.
\newblock The geometry of truth: Emergent linear structure in large language model representations of true/false datasets.
\newblock \emph{arXiv preprint arXiv:2310.06824}.

\bibitem[{Meng et~al.(2022)Meng, Bau, Andonian, and Belinkov}]{meng2022locating}
Kevin Meng, David Bau, Alex Andonian, and Yonatan Belinkov. 2022.
\newblock Locating and editing factual associations in {GPT}.
\newblock In \emph{Proceedings of NeurIPS}.

\bibitem[{M{\"u}ndler et~al.(2024)M{\"u}ndler, He, Jenko, and Vechev}]{mundler2024selfcontradiction}
Niels M{\"u}ndler, Jingxuan He, Slobodan Jenko, and Martin Vechev. 2024.
\newblock Self-contradictory hallucinations of large language models: Evaluation, detection and mitigation.
\newblock In \emph{Proceedings of ICLR}.

\bibitem[{{NVIDIA}(2024)}]{nvidia2024nemotron}
{NVIDIA}. 2024.
\newblock Nemotron-4 technical report.
\newblock \emph{arXiv preprint arXiv:2406.11704}.

\bibitem[{Pal et~al.(2024)}]{pal2024medhallu}
Vibhor Pal and 1 others. 2024.
\newblock {MedHalu}: Hallucinations in responses to healthcare queries by large language models.
\newblock \emph{arXiv preprint}.

\bibitem[{Qin et~al.(2025)}]{qin2025llmcas}
Yuheng Qin and 1 others. 2025.
\newblock Achilles' heel: Identifying and leveraging critical neurons for {LLM} hallucination correction via hierarchical {RL}.
\newblock \emph{arXiv preprint arXiv:2510.10238}.

\bibitem[{{Qwen Team}(2025)}]{qwen2025qwen25}
{Qwen Team}. 2025.
\newblock Qwen2.5 technical report.
\newblock \emph{arXiv preprint arXiv:2412.15115}.

\bibitem[{Rawte et~al.(2023)Rawte, Sheth, and Das}]{rawte2023survey}
Vipula Rawte, Amit Sheth, and Amitava Das. 2023.
\newblock A survey of hallucination in ``large'' foundation models.
\newblock \emph{arXiv preprint arXiv:2309.05922}.

\bibitem[{Su et~al.(2025)}]{su2025value}
Hangyu Su and 1 others. 2025.
\newblock Value neurons: Discovering and locating value representations in language models.
\newblock In \emph{Findings of EMNLP}.

\bibitem[{Tian et~al.(2024)}]{tian2024codehalu}
Yuchen Tian and 1 others. 2024.
\newblock {CodeHalu}: Code hallucinations in {LLMs} driven by execution-based verification.
\newblock \emph{arXiv preprint}.

\bibitem[{Turpin et~al.(2024)Turpin, Michael, Perez, and Bowman}]{turpin2024language}
Miles Turpin, Julian Michael, Ethan Perez, and Samuel~R. Bowman. 2024.
\newblock Language models don't always say what they think: Unfaithful explanations in chain-of-thought prompting.
\newblock \emph{Proceedings of NeurIPS}.

\bibitem[{Wei et~al.(2022)Wei, Wang, Schuurmans, Bosma, Ichter, Xia, Chi, Le, and Zhou}]{wei2022chain}
Jason Wei, Xuezhi Wang, Dale Schuurmans, Maarten Bosma, Brian Ichter, Fei Xia, Ed~Chi, Quoc Le, and Denny Zhou. 2022.
\newblock Chain-of-thought prompting elicits reasoning in large language models.
\newblock In \emph{Proceedings of NeurIPS}.

\bibitem[{Yu and Ananiadou(2024)}]{yu2024neuronlevel}
Zeping Yu and Sophia Ananiadou. 2024.
\newblock Neuron-level knowledge attribution in large language models.
\newblock In \emph{Proceedings of EMNLP}, pages 3267--3280.

\bibitem[{Zheng et~al.(2021)Zheng, Guha, Anderson, Henderson, and Ho}]{zheng2021casehold}
Lucia Zheng, Neel Guha, Brandon~R. Anderson, Peter Henderson, and Daniel~E. Ho. 2021.
\newblock When does pretraining help? assessing self-supervised learning for law and the {CaseHOLD} dataset of 53,137+ legal holdings.
\newblock In \emph{Proceedings of the International Conference on AI and Law}.

\bibitem[{Zhou et~al.(2019)Zhou, Liu, Siow, Du, and Liu}]{zhou2019devign}
Yaqin Zhou, Shangqing Liu, Jingkai Siow, Xiaoning Du, and Yang Liu. 2019.
\newblock Devign: Effective vulnerability identification by learning comprehensive program semantics via graph neural networks.
\newblock In \emph{Proceedings of NeurIPS}.

\end{thebibliography}
\end{document}